\documentclass[sigconf,natbib=false]{acmart}
\AtBeginDocument{%
  }

\renewcommand\footnotetextcopyrightpermission[1]{}

\RequirePackage[
  datamodel=acmdatamodel,
  style=acmnumeric,
  sorting=none,
]{biblatex}

\usepackage{amssymb}
\usepackage{multirow}
\usepackage{array}
\usepackage[table]{xcolor}   
\usepackage{tcolorbox}
\definecolor{HeaderGray}{RGB}{235,235,235}
\usepackage{makecell}
\usepackage{algorithmic}

\usepackage{soul} 
\sethlcolor{red!6}

\usepackage{pifont}
\usepackage{enumitem}
\definecolor{LineGray}{gray}{0.6} 
\usepackage{arydshln}
\usepackage{tabularx}
\newcolumntype{Y}{>{\centering\arraybackslash}X}

\acmSubmissionID{6271}

\begin{document}

\title{Cross-Resolution Semantic Transfer for Robust Text-to-Image Retrieval in Low-Resolution Surveillance}

\author{Wenjie Qian$^a$, Bin Yang$^b$, Xiao Wang$^{a,c}$, Wenke Huang$^b$, Ling Mei$^{a,c}$, Xin Xu$^{a,c}$, Mang Ye$^b$}
\affiliation{
\institution{$^a$School of Computer Science and Technology, Wuhan University of Science and Technology\\ $^b$School of Computer Science, National Engineering Research Center for Multimedia Software, Wuhan University\\$^c$Hubei Province Key Laboratory of Intelligent Information Processing and Real-time Industrial System, Wuhan University of Science and Technology
}
\city{}
\country{}
}








\renewcommand{\shortauthors}{Trovato et al.}

\begin{abstract}
Text-to-image person re-identification (TIPR) retrieves target persons using natural language descriptions. 
However, existing methods largely overlook resolution variance in real-world surveillance. They characterize cross-resolution TIPR through two coupled failure modes: \textit{Evidence Reliability Collapse (ERC)}, where degraded visual tokens become unreliable for grounding fine-grained text, and \textit{Ranking Distribution Drift (RDD)}, where mixed-resolution galleries distort similarity neighborhoods and destabilize retrieval rankings.
To address this challenge, we propose Cross-Resolution Semantic Transfer (CRST), a CLIP-style framework with three modules: resolution-conditioned reasoning, text-guided refinement and CR-RDA. Resolution-conditioned reasoning estimates token reliability to suppress corrupted evidence. Text-guided refinement injects semantic priors to recover discriminative cues. CR-RDA transfers HR neighborhood geometry to stabilize LR ranking under mixed resolutions.
Experiments on CUHK-PEDES, ICFG-PEDES, and RSTPReid show that CRST improves ultra-low-resolution Rank-1 and mAP on average by 5.7\% and 5.3\%, while stabilizing mixed-resolution retrieval without sacrificing high-resolution accuracy.
\textit{\textcolor{magenta}{
The code will be made publicly available.}}
\end{abstract}



\received{20 February 2007}
\received[revised]{12 March 2009}
\received[accepted]{5 June 2009}

\maketitle

\section{Introduction}
Text-to-image person re-identification (TIPR) has become a key technique for intelligent surveillance~\cite{bib1,bib1-1}, enabling users to search target persons in large-scale galleries using natural language descriptions~\cite{bib51}.
Benefiting from the rapid advancement of vision-language models such as CLIP~\cite{bib2}, recent methods have achieved strong performance by learning fine-grained cross-modal alignments between visual attributes and textual semantics~\cite{bib3}.
Despite this progress, real-world surveillance departs from the idealized assumption of high-quality imagery: captured images often exhibit drastic resolution variance due to camera distance, hardware limitations, and transmission constraints, making low-resolution (LR) retrieval an unavoidable open-world challenge.

\begin{figure}[t]
  \centering
  \includegraphics[width=0.48\textwidth]{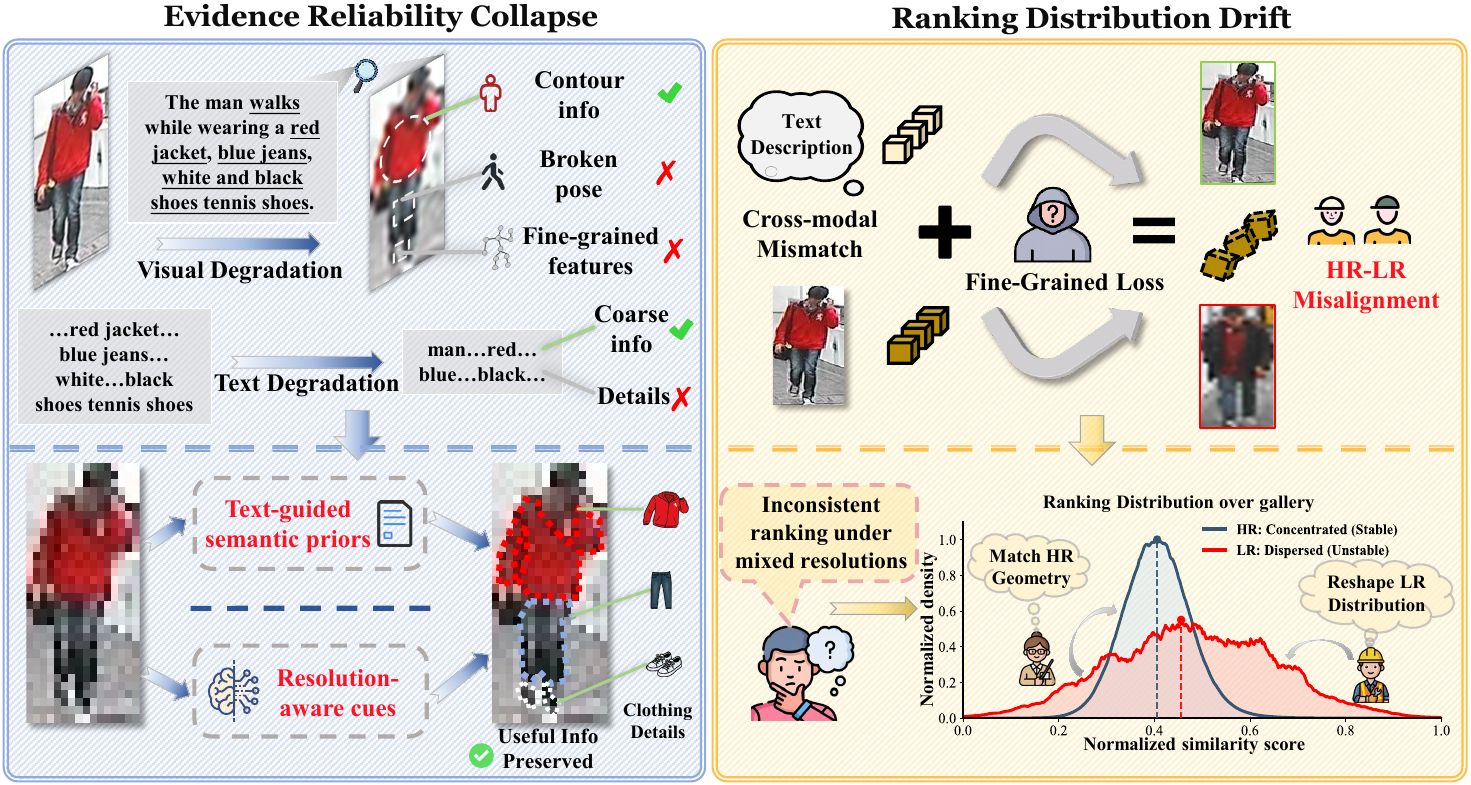}
  \caption{{\textbf{Illustration of motivation.} Problem \textbf{\textit{\hl{\strut I) Evidence Reliability Collapse (ERC)\strut}}}:} Resolution degradation corrupts fine-grained evidence, causing cross-modal mismatch and incorrect top-ranked retrieval. Problem \textbf{\textit{\hl{\strut II) Ranking Distribution Drift (RDD)\strut}}}: Mixed-resolution galleries distort similarity ordering, leading to ranking inconsistency across resolutions.}
  \label{fig:problem_show}
\end{figure}

Resolution variance is a pervasive yet under-modeled factor in real-world TIPR. 
Most existing TIPR frameworks are mainly developed and evaluated under standard HR or unified-resolution settings, where the retrieval space is implicitly assumed to be resolution-homogeneous. 
In real surveillance, however, images of different resolutions often coexist due to camera distance, hardware limitations, and transmission constraints, leading to two coupled challenges. 
First, low resolution weakens or removes fine-grained cues described by text, making cross-modal grounding unreliable. 
Second, when HR and LR samples coexist in the same gallery, the similarity landscape is reshaped, causing ranking instability across resolutions. 
As illustrated in Figure~\ref{fig:problem_show}, these effects make a baseline without explicit cross-resolution modeling prone to noisy grounding and incorrect retrieval.
This observation exposes problem \hl{\textit{\textbf{I) \textbf{\textit{Evidence Reliability Collapse}}:} how can we perform reliable fine-grained cross-modal grounding when resolution degradation corrupts visual evidence and induces severe text--image mismatch?}}

Mixed-resolution galleries introduce a practical failure mode that is largely overlooked by current TIPR systems. 
Most existing TIPR objectives are optimized and validated under resolution-homogeneous protocols, implicitly assuming a unified-resolution retrieval space. 
In deployed surveillance, however, HR and LR samples commonly coexist and compete in the same gallery. 
Such resolution heterogeneity reshapes the similarity landscape by compressing inter-instance margins and amplifying ranking instability, especially when low-resolution samples provide weaker and less reliable evidence~\cite{bib7-1,bib8}. 
As a result, the top-ranked retrieval results can become inconsistent and sensitive to resolution composition, even for the same query and identity.
This motivates problem \hl{\textit{\textbf{II) \textbf{\textit{Ranking Distribution Drift}}:} how can we maintain stable retrieval rankings when different-resolution samples coexist and distort similarity ordering across resolutions?}}

To address the above problems, we take a semantic-transfer view.
Rather than hallucinating pixels via separate super-resolution (SR) stages~\cite{bib5}, our goal is to make LR retrieval behave like its HR counterpart in both cross-modal alignment and ranking behavior.
We propose \textbf{Cross-Resolution Semantic Transfer (CRST)}, a unified CLIP-style framework that enforces semantic transfer across resolutions under a paired dual training protocol.
For problem {\setlength{\fboxsep}{1pt}\colorbox{red!7}{\textit{\textbf{I)}}}}, CRST models resolution-dependent evidence reliability via a \textit{\textbf{Resolution-Conditioned Reasoner (RCR)}} to estimate token-wise trustworthiness for reliability-aware reasoning, and further introduces a \textit{\textbf{Text-Guided Refiner (TGR)}} to inject modality-invariant semantic priors that recover discriminative cues in the embedding space.
For problem {\setlength{\fboxsep}{1pt}\colorbox{red!7}{\textit{\textbf{II)}}}}, CRST proposes \textit{\textbf{Cross-Resolution Ranking Distribution Alignment (CR-RDA)}} to transfer HR neighborhood topology to LR retrieval, stabilizing rankings under mixed-resolution galleries.
We summarize our contributions as follows:
\begin{enumerate}[label=\large\ding{\numexpr181+\arabic*\relax}, leftmargin=*, labelsep=0.5em]
    \item \textit{\textbf{Cross-Resolution TIPR.}} We formulate a cross-resolution TIPR setting for the practical yet under-explored case where HR and LR samples coexist in the same retrieval space, and characterize its main challenges through two coupled failure modes: ERC and RDD.
    
    \item \textit{\textbf{Semantic Transfer Framework: CRST.}} We present CRST, a unified CLIP-style semantic transfer framework that enables LR retrieval to inherit reliable evidence, discriminative semantics, and stable ranking geometry from HR, via resolution-conditioned token reasoning, text-guided refinement, and cross-resolution ranking distribution alignment.

    \item \textit{\textbf{Real-World Usability Validation.}} We verify CRST on three public benchmarks under standard HR as well as cross-resolution protocols. CRST yields clear robustness gains over strong CLIP-based baselines. Averaged across three benchmarks, Ultra-LR performance increases by 5.7\% in R@1 and 5.3\% in mAP, while maintaining competitive HR accuracy.
\end{enumerate}

\section{Related Work}
\subsection{Text-to-Image Person Retrieval}
Text-to-image person retrieval aims to retrieve the person-of-interest from a large-scale image gallery using natural-language descriptions by learning a shared vision--language embedding space~\cite{bib9,bib47}.
Early works typically adopted dual encoders with global matching objectives, while later methods enhanced fine-grained alignment through part and region representations, phrase-aware matching, or attention-based interaction~\cite{bib10}. 
With the advance of vision--language pre-training, CLIP-style dual encoders provide stronger cross-modal representations and have become a mainstream backbone for TIPR~\cite{bib13,bib56}. 
Building upon these backbones, recent approaches further improve interaction and alignment via relation modeling and distribution-level constraints that stabilize global ranking.  
Recent CLIP-based TIPR frameworks enhance fine-grained alignment via token-level interaction and distribution-level ranking regularization, achieving strong performance under standard HR protocols. 
However, they are mainly developed and evaluated under standard HR or unified-resolution settings, without explicitly addressing mixed-resolution galleries or the loss of fine-grained visual evidence under severe low resolution.
\begin{figure*}[tb]
  \centering
  \includegraphics[width=1.0\textwidth]{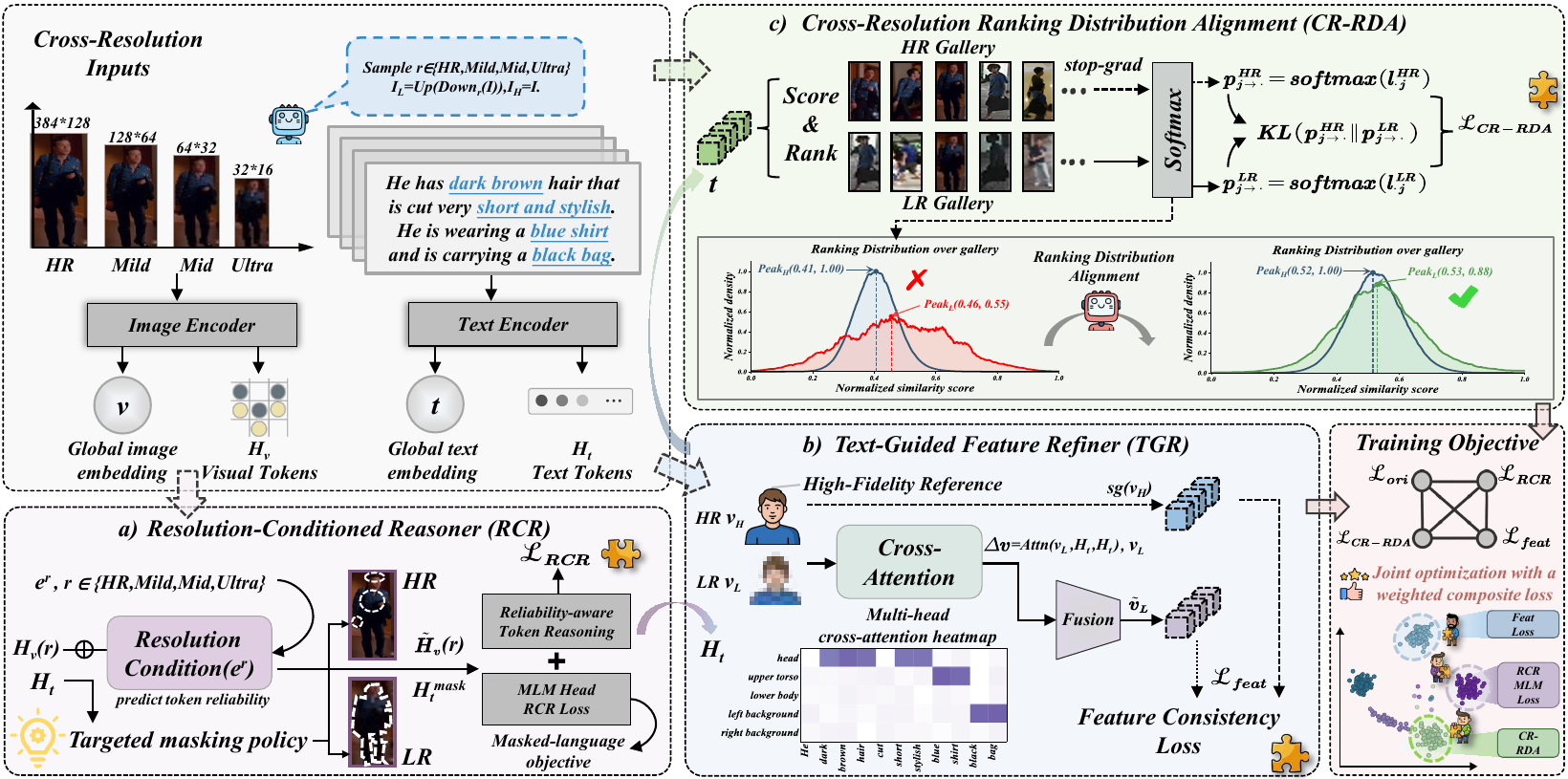}
  \caption{\textbf{The pipeline of the CRST. } 
  CRST mitigates ERC and RDD by enforcing HR-referenced robustness constraints on paired dual views:
  (a) RCR performs reliability-aware evidence reasoning by estimating resolution-conditioned token reliability to suppress corrupted tokens;
  (b) TGR injects text-guided semantic priors to correct degraded embeddings and strengthen fine-grained alignment;
  (c) CR-RDA enforces HR-referenced neighborhood-geometry consistency to stabilize hard-negative ordering under mixed and Ultra-LR retrieval.
  }
  \label{fig:crst}
\end{figure*}

\subsection{Cross-Resolution Retrieval and Alignment}
Cross-resolution recognition and retrieval have been widely studied in image ReID and general retrieval~\cite{bib48}, where heterogeneous resolutions induce substantial detail loss and distribution shifts~\cite{bib14,bib60}. 
Existing solutions mainly follow two paradigms: restoration-based pipelines that recover high-frequency details before matching~\cite{bib15} and resolution-invariant representation learning that enforces cross-scale consistency via feature and similarity consistency constraints~\cite{bib16,bib61}. 
Restoration is often computationally expensive and may introduce artifacts, whereas purely invariant learning can overlook that the reliability of visual cues varies with resolution. 
Moreover, most cross-resolution studies focus on image-to-image matching and do not directly address TIPR, where language remains fine-grained but visual evidence becomes ambiguous at low resolution, amplifying cross-modal misalignment~\cite{bib17,bib57}.
Although auxiliary guidance and reference-guided regularization have been explored for robust cross-modal retrieval, few works explicitly formulate cross-resolution TIPR as a semantic transfer problem, transferring trustworthy evidence~\cite{bib62}, discriminative semantics, and ranking structure from HR to LR within a single end-to-end framework~\cite{bib18,bib58}. 
Therefore, different from prior paradigms, we propose CRST, which treats HR as a stop-gradient reference and enforces robustness constraints on token-level evidence reliability, feature-space semantic correction, and hard-negative ordering consistency, enabling stable retrieval under Ultra-LR and mixed-resolution galleries in a CLIP-style framework.

\section{Methodology}
\subsection{Problem Setting and Preliminaries}
We study cross-resolution text-to-image person retrieval.
Each training sample contains a text--image pair $(T,I)$ with identity label $y$, and the image has a resolution level $r\in\mathcal{R}$.
The text encoder outputs token features $\mathbf{H}^t$ and an embedding $\mathbf{t}\in\mathbb{R}^d$, while the image encoder outputs resolution-conditioned visual tokens $\mathbf{H}^v(r)$ and an embedding $\mathbf{v}\in\mathbb{R}^d$.
Cross-modal similarity is computed by
\begin{equation}
\ell(\mathbf{t},\mathbf{v})=\frac{1}{\tau}\cdot
\frac{\mathbf{t}^{\top}\mathbf{v}}{\|\mathbf{t}\|\|\mathbf{v}\|}.
\label{eq:1}
\end{equation}
The cosine term measures cross-modal similarity, and $\ell(\cdot)$ denotes the temperature-scaled retrieval score.

\textbf{Resolution-aware visual tokens. }
To make the encoder aware of quality, we model resolution as a categorical condition via a learnable embedding $\mathbf{e}_r \in \mathbb{R}^d$, where {$r \in \mathcal{R}$} denotes the resolution level.
Let $\{\mathbf{x}_i\}_{i=0}^L$ denote the token embeddings before the visual transformer (with $L+1$ visual tokens). We inject resolution by~\cite{bib19}
\begin{equation}
\begin{gathered}
\mathbf{x}^{(r)}_i=\mathbf{x}_i+\mathbf{e}_r,\; i=0,\ldots,L, \\ \mathbf{H}^v(r)=\mathrm{ViT}(\{\mathbf{x}^{(r)}_i\}_{i=0}^{L}).
\end{gathered}
\label{eq:2}
\end{equation}
Thus the resulting global embedding $\mathbf{v}=F_I(I,r) \in \mathbb{R}^d$ becomes resolution-conditioned, enabling downstream cross-modal reasoning to adapt to resolution variance.
We emphasize that $\mathbf{e}_r$ only provides a global resolution condition. The core of RCR is to predict token-wise evidence reliability conditioned on $r$ and use it to modulate attention, rather than uniformly shifting all tokens.

\textbf{Resolution-Contrasting Dual Views. }
For each image, we construct HR and LR views $I^H$ and $I^L$ sharing $(T,y)$.
HR branch serves as a stop-gradient reference, while the LR branch learns robustness via refinement and ranking-geometry alignment~\cite{bib20}.

\subsection{Overview of CRST Framework}
As shown in Figure~\ref{fig:crst}, CRST formulates robustness as HR-referenced semantic consistency between HR/LR views.
To address the two coupled failure, CRST is organized around three robustness constraints: i) \textit{reliability-aware evidence selection} to avoid grounding on corrupted LR tokens; ii) \textit{prior-guided semantic correction} to recover cues in embedding space; iii) \textit{hard-negative ordering consistency} to stabilize rankings under mixed-resolution galleries.
We operationalize these constraints as RCR, TGR, and CR-RDA within a CLIP dual encoder trained on dual views.
All constraints are applied during training; inference reduces to Eq.~(\ref{eq:1}) with zero overhead.

\subsection{Resolution-Conditioned Reasoner}
We introduce RCR as a resolution-conditioned token-level reasoner trained with an RCR objective. 
ERC does not corrupt all visual tokens uniformly: even at the same resolution, some tokens remain informative while others become unreliable due to blur and fine-grained detail loss~\cite{bib59}. 
We therefore model evidence reliability as resolution-aware but token-specific. 
Given resolution-conditioned visual tokens {\footnotesize $\mathbf{H}^v(r)=\{\mathbf{h}^v_k(r)\}_{k=0}^{L}$} and the resolution embedding $\mathbf{e}_r$, we estimate for each token a reliability gate
\begin{equation}
\begin{gathered}
\rho_k=\sigma\!\left(\mathrm{MLP}\left([\mathrm{LN}(\mathbf{h}^v_k(r));\mathbf{e}_r]\right)\right),\\ \qquad
\tilde{\mathbf{h}}^v_k(r)=\rho_k \cdot \mathbf{h}^v_k(r),
\end{gathered}
\label{eq:rcr_rho}
\end{equation}
where $\rho_k\in(0,1)$ measures token trustworthiness for cross-modal grounding under resolution level $r$. $\mathbf{h}^v_k(r)$ captures token-specific content, $\mathbf{e}_r$ provides global degradation context. We use scalar gating to modulate evidence contribution before cross-modal attention, suppressing unreliable tokens and retaining informative ones.

Let $T^{\text{mask}}$ denote the masked text sequence and $\mathbf{H}^t_{\text{mask}}$ the corresponding text token embeddings.
RCR performs cross-attention from masked text tokens to the gated visual tokens:
\begin{equation}
\begin{gathered}
\mathbf{Z}^t=\mathrm{Attn}(\mathbf{H}^t_{\text{mask}},\tilde{\mathbf{H}}^v(r),\tilde{\mathbf{H}}^v(r)).
\end{gathered}
\label{eq:3}
\end{equation}
Masked grounding supervision makes $\rho_k$ learn to suppress corrupted evidence and select reliable tokens for cross-modal reasoning under resolution degradation.
The resulting objective is:
\begin{equation}
\begin{gathered}
\mathcal{L}_{\text{RCR}} = -\sum_{i \in \mathcal{M}} \log p_{\theta}\left( w_i \mid T^\text{mask}, I(r) \right).
\end{gathered}
\label{eq:4}
\end{equation}
\textbf{Resolution-conditioned masking. }
A major difficulty in cross-resolution TIPR is that fine-grained textual semantics remain intact, but fine-grained visual evidence vanishes in LR images. Uniform masking may force the model to explain fine-grained attributes from corrupted LR evidence, amplifying misalignment.
We therefore sample masked positions with a resolution-dependent policy $M \sim \pi_r$.
We divide tokens into two groups: attribute tokens describing fine-grained appearance cues and coarse tokens describing higher-level semantics.
The masking ratio is shifted from attribute-heavy in HR to coarse-heavy in LR (HR: 70/30, Mild-LR: 60/40, Mid-LR: 45/55, Ultra-LR: 30/70 for attribute/coarse tokens), so that supervision matches the reliability of visual evidence at each resolution.
Full definitions, formulas, and pseudocode are provided in Appendix~A.

\subsection{Text-Guided Feature Refinement}
LR images may lose discriminative cues explicitly present in language. 
TGR treats text as a semantic prior and performs feature-space correction on LR embeddings, narrowing the HR-LR gap when fine-grained evidence vanishes~\cite{bib49,bib53}.
Let $\mathbf{v}^L$ be the LR global image embedding and $\mathbf{H}^t$ the text token embeddings. 
TGR forms a prior-guided correction by letting the LR embedding query discriminative text tokens.
We instantiate it with a lightweight cross-modal attention operator where $\mathbf{v}^L$ serves as the query and $\mathbf{H}^t$ as keys/values:
\begin{equation}
\begin{gathered}
\Delta\mathbf{v} = \text{Attn}\left( \mathbf{v}^L, \mathbf{H}^t, \mathbf{H}^t \right), \\ \quad g = \sigma\left( \mathbf{w}^\top [\mathbf{v}^L; \Delta\mathbf{v}] \right), \quad \tilde{\mathbf{v}}^L = \text{LN}\left( \mathbf{v}^L + g \cdot \Delta\mathbf{v} \right),
\end{gathered}
\label{eq:5}
\end{equation}
here $g$ adaptively controls the correction strength, preventing over-reliance on text when visual evidence is reliable, the update is a retrieval-oriented feature correction rather than a generic residual adapter~\cite{bib22}.
This design restores discriminative semantics directly in the retrieval feature space, avoiding pixel-level restoration artifacts while improving cross-modal alignment.

We further impose HR-referenced feature consistency using the HR embedding $\mathbf{v}^H$:
\begin{equation}
\begin{gathered}
\mathcal{L}_{\text{feat}} = \left\| \tilde{\mathbf{v}}^L - \text{sg}\left( \mathbf{v}^H \right) \right\|_2^2,
\end{gathered}
\label{eq:6}
\end{equation}
where $\mathrm{sg}(\cdot)$ keeps the HR reference stable, preventing LR noise from drifting the alignment target.

\subsection{Label-based Similarity Distribution Matching}
Given a mini-batch of $N$ text--image pairs with identity labels $\{y_i\}_{i=1}^N$,
we define the label-induced target distribution for text-to-image retrieval as
\begin{equation}
\begin{array}{c}
q^{\text{gt}}_{j\rightarrow i}=
\begin{cases}
\frac{1}{|\mathcal{P}(j)|}, & y_i = y_j,\\
0, & \text{otherwise},
\end{cases} \\[3mm]
\mathcal{P}(j)=\{i \mid y_i=y_j\},
\end{array}
\label{eq:7}
\end{equation}
and the predicted similarity distribution as $p_{j\rightarrow i}=\mathrm{softmax}(\ell_{ij})$, where $\ell_{ij}= s(\mathbf{t}_j,\mathbf{v}_i)/\tau$.
Similarity Distribution Matching (SDM) optimizes supervised retrieval by matching $p$ to $q^{\text{gt}}$:
\begin{equation}
\begin{array}{c}
\mathcal{L}_{\mathrm{SDM}}^{T\rightarrow I}=\frac{1}{N}\sum_{j=1}^N \mathrm{KL}\!\left(q^{\text{gt}}_{j\rightarrow \cdot}\,\|\,p_{j\rightarrow \cdot}\right),\\[3mm]
\mathcal{L}_{\mathrm{SDM}}=\mathcal{L}_{\mathrm{SDM}}^{T\rightarrow I}+\mathcal{L}_{\mathrm{SDM}}^{I\rightarrow T}.
\end{array}
\label{eq:8}
\end{equation}
Since $q^{\text{gt}}$ assigns zero mass to negatives, $\mathcal{L}_{\mathrm{SDM}}$ is invariant to permutations within negatives and thus cannot supervise hard-negative ordering, causing near-neighbor rank swaps under mixed resolutions.
We next introduce CR-RDA to explicitly regularize hard-negative ordering via HR-referenced ranking-geometry transfer.

\subsection{Cross-Resolution Ranking Distribution Alignment}
Feature refinement alone cannot guarantee stable rankings under severe degradation, because retrieval depends on the relative ordering among competing candidates rather than on a single similarity score. 
We therefore transfer HR neighborhood topology by regularizing the LR ranking distribution toward an HR high-fidelity reference distribution~\cite{bib24}. 
Given a minibatch of $N$ paired samples, we compute cosine logits $\ell_{ij}$ and define the text-to-image matching distribution:
\begin{equation}
\begin{gathered}
p_{j\rightarrow i}=\frac{\exp(\ell_{ij})}{\sum_{k=1}^{N}\exp(\ell_{kj})}.
\end{gathered}
\label{eq:9}
\end{equation}
where $\mathbf{v}_i$ and $\mathbf{t}_j$ denote the image and text embeddings of the $i$-th image and $j$-th text in the mini-batch. 
Thus, each query induces a normalized ranking distribution over batch candidates, in which the probability mass concentrates on more competitive neighbors and the relative probabilities encode local ordering and similarity margins. This distribution therefore provides a differentiable representation of retrieval neighborhood geometry for subsequent HR-to-LR alignment.

For geometry transfer, we compute the HR distribution using the HR embedding $\mathbf{v}^H$ and the LR distribution using the refined LR embedding $\tilde{\mathbf{v}}^L$. 
We stop-gradient the HR reference branch to keep a reference, preventing the alignment target from being corrupted by LR-induced ranking noise, i.e., $\ell^{H}_{ij}= s(\mathbf{t}_j,\mathrm{sg}(\mathbf{v}^H_i))/\tau$ and $p^{H}_{j\rightarrow\cdot}=\mathrm{softmax}(\ell^{H}_{\cdot j})$, while the LR branch uses $\ell^{L}_{ij}= s(\mathbf{t}_j,\tilde{\mathbf{v}}^L_i)/\tau$ and $p^{L}_{j\rightarrow\cdot}=\mathrm{softmax}(\ell^{L}_{\cdot j})$.
We enforce HR-referenced ranking-geometry consistency by minimizing KL$(p^H_{j\rightarrow\cdot}\,\|\,p^L_{j\rightarrow\cdot})$:
\begin{equation}
\begin{gathered}
\mathcal{L}_{\text{CR-RDA}}^{T \to I} = \frac{1}{N} \sum_{j=1}^N \text{KL}\left( p_{j \to \cdot}^H \mid\mid p_{j \to \cdot}^L \right).
\end{gathered}
\label{eq:10}
\end{equation}
Since the HR reference branch is stop-gradient, minimizing Eq.~(\ref{eq:10}) yields $\partial \mathcal{L}/\partial \ell_{ij}^L = p_{j \to i}^L - p_{j \to i}^H$, which directly corrects each candidate's relative probability and preserves hard-negative ordering.

We apply the same idea symmetrically for image-to-text matching. Define
\begin{equation}
\begin{gathered}
p_{i\rightarrow j}=\frac{\exp(\ell_{ij})}{\sum_{k=1}^{N}\exp(\ell_{ik})},
\end{gathered}
\label{eq:11}
\end{equation}
and compute $\mathcal{L}_{\text{CR-RDA}}^{I \to T}$ analogously. The final CR-RDA loss:
\begin{equation}
\begin{gathered}
\mathcal{L}_{\text{CR-RDA}} = \mathcal{L}_{\text{CR-RDA}}^{I \to T} + \mathcal{L}_{\text{CR-RDA}}^{T \to I}.
\end{gathered}
\label{eq:12}
\end{equation}
Unlike positive-only supervision that leaves negatives unconstrained, CR-RDA provides neighborhood supervision over in-batch candidates, penalizing near-neighbor rank swaps and preserving hard-negative ordering under mixed and Ultra-LR settings~\cite{bib26}.

\begin{table*}[t]
    \centering
    \scriptsize
    \definecolor{HeaderGray}{gray}{0.92}
    \definecolor{BodyGrayOdd}{gray}{0.95}
    \definecolor{BodyGrayEven}{gray}{1.0}
    \definecolor{OursBlue}{HTML}{E0F7FA}
    \definecolor{LineGray}{gray}{0.6}
    \setlength{\dashlinedash}{6pt}
    \setlength{\dashlinegap}{2.2pt}
    \setlength{\doublerulesep}{1.2pt}
    \doublerulesepcolor{white}
    \newcommand{\DatasetRule}{\vrule width 0.8pt}
    \newcommand{\ThickDashLine}{\hline}
    \resizebox{\textwidth}{!}{
        \renewcommand{\arraystretch}{1.15}
        \rowcolors{5}{BodyGrayOdd}{BodyGrayEven}

        \begin{tabular}{ c || cc cc !{\DatasetRule} cc cc !{\DatasetRule} cc cc }
            \hline
            \rowcolor{HeaderGray}
             & \multicolumn{4}{c !{\DatasetRule}}{\textbf{CUHK-PEDES}}
             & \multicolumn{4}{c !{\DatasetRule}}{\textbf{ICFG-PEDES}}
             & \multicolumn{4}{c}{\textbf{RSTPReid}} \\
            \cline{2-13}
            \rowcolor{HeaderGray}
             & \multicolumn{2}{c}{Ultra-LR} & \multicolumn{2}{c !{\DatasetRule}}{Mixed}
             & \multicolumn{2}{c}{Ultra-LR} & \multicolumn{2}{c !{\DatasetRule}}{Mixed}
             & \multicolumn{2}{c}{Ultra-LR} & \multicolumn{2}{c}{Mixed} \\
            \cline{2-13}
            \rowcolor{HeaderGray}
            \multirow{-3}{*}{Methods}
            & R@1 & mAP & R@1 & mAP
            & R@1 & mAP & R@1 & mAP
            & R@1 & mAP & R@1 & mAP \\
            \ThickDashLine
            \rowcolor{white}
            \multicolumn{13}{l}{\hspace{0.6em}\textit{\color{LineGray}{Protocol TD: original training}}} \\
            \ThickDashLine
            CFine~\cite{bib36}        & 51.93 & 45.50 & 63.67 & 55.56 & 38.47 & 19.11 & 53.33 & 29.24 & 36.12 & 27.05 & 45.31 & 34.38 \\
            UniPT~\cite{bib35}        & 51.31 & 45.01 & 62.86 & 54.89 & 37.84 & 18.72 & 52.30 & 28.91 & 37.04 & 27.70 & 46.44 & 35.21 \\
            TBPS-CLIP~\cite{bib39}    & 54.97 & 48.12 & 67.28 & 58.76 & 40.18 & 20.27 & 55.02 & 30.64 & 44.24 & 33.12 & 55.50 & 41.93 \\
            MMRef~\cite{bib38-1}        & 54.22 & 48.36 & 66.27 & 58.77 & 40.06 & 19.99 & 55.20 & 30.26 & 40.31 & 30.26 & 50.45 & 38.18 \\
            DM-Adapter~\cite{bib38-2}   & 54.47 & 48.36 & 66.85 & 58.91 & 39.88 & 19.43 & 55.31 & 29.89 & 42.66 & 31.94 & 53.61 & 40.48 \\
            IRRA~\cite{bib7}         & 54.75 & 48.69 & 67.10 & 59.35 & 40.05 & 19.71 & 55.64 & 30.16 & 42.97 & 32.14 & 53.93 & 40.82 \\
            
            \ThickDashLine
            \rowcolor{white}
            \multicolumn{13}{l}{\hspace{0.6em}\textit{\color{LineGray}{Protocol PHL: paired HR/LR training}}} \\
            \ThickDashLine
            {DM-Adapter\dag} & 56.91 & 51.27 & 68.38 & 61.12 & 41.87 & 21.52 & 56.64 & 31.51 & 44.85 & 34.26 & 55.07 & 42.10 \\
            \hline
            \rowcolor{OursBlue}
            \textbf{CRST (Ours)} & \textbf{61.81} & \textbf{55.84} & \textbf{70.36} & \textbf{64.05}
                               & \textbf{46.17} & \textbf{25.94} & \textbf{58.11} & \textbf{34.45}
                               & \textbf{49.55} & \textbf{37.80} & \textbf{56.20} & \textbf{44.69} \\
            \hline
        \end{tabular}
    }
    \caption{\textbf{Cross-resolution robustness. }
    Comparison on CUHK-PEDES, ICFG-PEDES, and RSTPReid under Ultra-LR and Mixed settings. 
    \textit{TD}: models follow original recipes under test-time degradation. \textit{PHL}: models are trained with paired dual views.}
    \label{tab:lr_all}
    \vspace{-2mm}
\end{table*}

\begin{figure*}[t]
  \centering
  \includegraphics[width=0.95\textwidth]{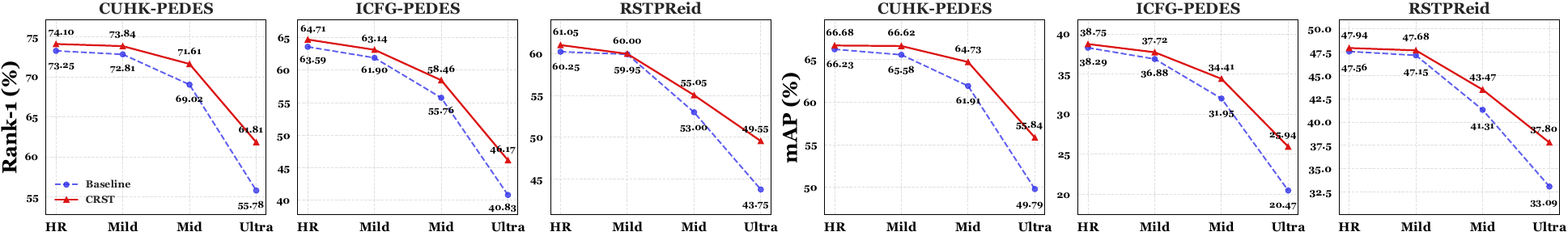}
  \caption{\textbf{Progressive degradation. }
  We compare baseline and CRST across four resolution settings, reporting Rank-1 and mAP.}
  \label{fig:rank_map_cmp}
  \vspace{-2mm}
\end{figure*}

\textbf{Difference between SDM and CR-RDA. }
SDM enforces retrieval correctness via positives, but is positive-only and permutation-invariant within negatives, so it cannot prevent near-neighbor rank swaps under resolution disturbance.
CR-RDA instead constrains a dense HR-referenced neighborhood geometry over candidates, explicitly supervising hard-negative ordering to mitigate RDD.

\textbf{Overall loss. } 
We optimize CRST with the sum of the RCR grounding loss, the text-guided refinement consistency, standard TIPR retrieval losses, and CR-RDA:
\begin{equation}
\begin{gathered}
\mathcal{L} = \mathcal{L}_{\text{RCR}} + \mathcal{L}_{\text{SDM}} + \lambda \mathcal{L}_{\text{CR-RDA}} + \beta \mathcal{L}_{\text{feat}} + \gamma \mathcal{L}_{\text{ID}}.
\end{gathered}
\label{eq:13}
\end{equation}
Here, $L_{SDM}$ is the standard similarity-distribution matching loss for retrieval correctness, 
$L_{CR\text{-}RDA}$ is the HR-referenced ranking-distribution alignment loss, and $L_{ID}$ is the identity classification loss. 
The coefficients $\lambda$, $\beta$, and $\gamma$ are the loss weights for $L_{CR\text{-}RDA}$, $L_{feat}$, and $L_{ID}$, respectively.

\subsection{Discussion}
Unlike existing TIPR methods mainly developed under standard HR or unified-resolution settings, CRST formulates text-to-image person retrieval as a cross-resolution problem. In real surveillance, low-resolution degradation weakens fine-grained evidence for cross-modal grounding, while HR/LR coexistence disrupts ranking stability in mixed-resolution galleries. By using HR as a semantic reference for evidence selection, feature refinement, ranking organization, CRST bridges this gap and achieves stronger Ultra-LR and Mixed robustness while preserving standard CLIP-style inference.

\begin{table*}[t]
    \centering
    \definecolor{HeaderGray}{gray}{0.92}
    \definecolor{BodyGrayOdd}{gray}{0.95}
    \definecolor{BodyGrayEven}{gray}{1.0}
    \definecolor{OursBlue}{HTML}{E0F7FA}

    \renewcommand{\arraystretch}{1.15} 
    \rowcolors{3}{BodyGrayOdd}{BodyGrayEven}

    \begin{tabularx}{\textwidth}{ >{\centering\arraybackslash}p{5.0cm} | >{\centering\arraybackslash}p{1.7cm} || YY | YY | YY }
        \hline
        \rowcolor{HeaderGray} 
         & & \multicolumn{2}{c|}{\textbf{CUHK-PEDES}} & \multicolumn{2}{c|}{\textbf{ICFG-PEDES}} & \multicolumn{2}{c}{\textbf{RSTPReid}} \\
        \cline{3-8}
        \rowcolor{HeaderGray} 
        \multirow{-2}{*}{Method} & \multirow{-2}{*}{Ref} & R@1 & mAP & R@1 & mAP & R@1 & mAP \\
        \hline
        CMPM/C~\cite{bib30} & ECCV18 & - & - & 43.51 & - & - & - \\
        TIMAM~\cite{bib31} & ICCV19 & 54.51 & - & - & - & - & - \\
        NAFS~\cite{bib32} & arXiv21 & 59.36 & 54.07 & - & - & - & - \\
        LBUL~\cite{bib44} & MM22 & 64.04& - & - & - & 45.55 & - \\
        IVT~\cite{bib45} & arXiv22 & 65.59 & 60.66 & 56.04 & - & 46.70 & - \\
        AXM-Net~\cite{bib33} & MM22 & 64.44 & 58.73 & - & - & - & - \\
        ISANet~\cite{bib42} & arXiv23 & 63.92 & - & 57.73 & - & - & - \\
        SSAN~\cite{bib28} & arXiv21 & 61.37 & - & 54.23 & - & 43.50 & - \\
        LC$R^2$S~\cite{bib29} & MM23 & 67.36 & 59.24 & 57.93 & 38.21 & 54.95 & 40.92 \\
        PBSL~\cite{bib40} & MM23 & 65.32 & - & 57.84 & - & 47.80 & - \\
        UniPT~\cite{bib35} & ICCV23 & 68.50 & - & 60.09 & - & 51.85 & - \\
        TP-TPS~\cite{bib37} & arXiv23 & 70.16 & \underline{66.32} & 60.64 & 42.78 & 50.65 & 43.11 \\
        CFine~\cite{bib36} & TIP23 & 69.57 & - & 60.83 & - & 50.55 & - \\
        TBPS-CLIP~\cite{bib39} & AAAI24 & \underline{73.54} & 65.38 & \textbf{65.05} & \textbf{39.83} &  \underline{61.95} & \textbf{48.26} \\
        (Li et al., 2024)~\cite{bib38} & arXiv24 & 71.59 & 65.03 & 60.93 & 36.44 & - & - \\
        RDE~\cite{bib34} & CVPR24 & 71.33 & 63.50 & 63.76 & 37.38 & \textbf{63.85} & 47.67 \\
        MMRef~\cite{bib38-1} & TMM25 & 72.25 & 65.23 & 63.50 & - & 56.20 & - \\
        DM-Adapter~\cite{bib38-2} & AAAI25 & 72.17 & 64.33 & 62.64 & 36.50 & 60.00 & 47.37 \\
        IRRA~\cite{bib7} & CVPR23 & 73.38 & 66.13 & 63.46 & 38.06 & 60.20 & 47.17 \\
        \hline
        \rowcolor{OursBlue} 
        \textbf{CRST (Ours)} & - & \textbf{74.10} & \textbf{66.68} & \underline{64.71} & \underline{38.75} & 61.05 & \underline{47.94} \\
        \hline
    \end{tabularx}

    \caption{\textbf{Comparison under the Standard HR setting. }
    Comparison on CUHK-PEDES, ICFG-PEDES, and RSTPReid Under the Standard HR Setting. The Best and Second-Best Results are
    Highlighted in Bold and \underline{Underline}.}
    \label{tab:hr_all}
    \vspace{-2mm}
\end{table*}

\section{Experiments}\frenchspacing
In this section, we describe the experimental setup, covering datasets, protocols, and metrics. Then, we provide details, followed by main results, ablation analysis, and qualitative evaluations. Results are presented under settings including HR and LR conditions.

\subsection{Experimental Setup}
\hspace*{1em}\textbf{Datasets. }
We evaluate CRST on three public benchmarks: CUHK-PEDES~\cite{bib27}, ICFG-PEDES~\cite{bib28}, and RSTPReid~\cite{bib29}.
Each dataset follows standard train/val/test splits. Training uses image-text pairs, while validation and test evaluate retrieval.

\textbf{Cross-resolution Protocol. }
To study resolution variance, we construct four resolution levels $\mathcal{R}=\{\text{HR},\allowbreak\text{Mild-LR},\allowbreak\text{Mid-LR},\allowbreak\text{Ultra-LR}\}$ by downsampling images to predefined sizes (HR: $384\times128$, Mild-LR: $128\times64$, Mid-LR: $64\times32$, Ultra-LR: $32\times16$).
We use bicubic interpolation for downsampling and resize back to $384\times128$ before feeding into CLIP, so that resolution changes reflect information loss rather than input-size mismatch; this serves as a controlled protocol rather than assuming real surveillance quality falls into four exact bins.
Besides reporting each resolution, we introduce a Mixed setting where each gallery image is assigned a resolution level by uniform sampling over $\mathcal{R}$, while practical robustness to coupled degradations, noisy resolution labels, and an SR-aware alternative is further analyzed in Appendix~B--D.

\textbf{Evaluation Metrics. }
We treat text descriptions as queries and rank images by similarity, reporting Rank-$k$ accuracy and mean Average Precision (mAP).
In the paper, we focus on Rank-1 (R@1) and mAP~\cite{bib29-1} to reflect top-1 correctness and ranking quality.

\begin{table}[t]
  \centering
  \definecolor{HeaderGray}{gray}{0.92}
  \definecolor{BodyGrayOdd}{gray}{0.95} 
  \definecolor{BodyGrayEven}{gray}{1.0} 
  \definecolor{LineGray}{gray}{0.6} 

  \setlength{\tabcolsep}{6pt}
  \renewcommand{\arraystretch}{1.1} 
  
  \resizebox{0.98\linewidth}{!}{
    \begin{tabular}{
      >{\centering\arraybackslash}m{1.2cm}
      !{\color{LineGray}\vrule}
      >{\centering\arraybackslash}m{1.6cm}
      !{\color{LineGray}\vrule}
      >{\centering\arraybackslash}m{2.2cm}
      !{\color{LineGray}\vrule}
      >{\centering\arraybackslash}m{1.4cm}
    }
      \hline
      
      \rowcolor{HeaderGray}
      Method & Params (M) & \shortstack[c]{Inference\\Speed (img/s)} & FLOPs (G) \\
      \hline
      
      \rowcolor{BodyGrayEven} 
      IRRA & 194.5 & 315.4 & 19.5 \\
      
      \rowcolor{BodyGrayOdd} 
      Ours & 196.7 & 312.7 & 19.5 \\
      
      \hline
    \end{tabular}
  }
  \caption{\textbf{Model complexity and efficiency comparison. }
  Params denotes learnable parameters. FLOPs and Speed are measured under the same hardware and batch configuration.}
  \label{tab:complexity}
  \vspace{-4mm}
\end{table}

\subsection{Implementation Details}
\hspace*{1em}\textbf{Backbone and Training Pipeline. }
CRST adopts a CLIP dual-encoder backbone with ViT-B/16 as the visual encoder.
The visual encoder is a 12-layer Vision Transformer with patch size 16, stride size 16, and 12 attention heads, while the paired CLIP text transformer uses a context length of 77, hidden width 512, 12 layers, and 8 attention heads.
The shared image-text embedding dimension is 512, and all models are initialized from the official CLIP pretrained checkpoint.
We resize all images to $384\times128$.
The cross-modal interaction module uses hidden size 512, 8 attention heads, and depth 4.
For cross-resolution training, each sample is associated with an HR view and a degraded LR view, and LR generation is activated with probability $p=0.5$.

\begin{table*}[htbp]
  \centering
  \footnotesize
  \definecolor{HeaderGray}{gray}{0.92}
  \definecolor{BodyGrayOdd}{gray}{0.95}
  \definecolor{BodyGrayEven}{gray}{1.0}
  \definecolor{OursBlue}{HTML}{E0F7FA}
  \definecolor{LineGray}{gray}{0.6} 

  \newcommand{\DatasetRule}{\vrule width 1.4pt}

  \resizebox{\textwidth}{!}{
    \renewcommand{\arraystretch}{1.15}
    \setlength{\tabcolsep}{1.5pt}
    \rowcolors{4}{BodyGrayOdd}{BodyGrayEven}
    \begin{tabular}{c !{\color{LineGray}\vrule} c c c !{\color{LineGray}\vrule} c c c c c !{\color{LineGray}\vrule} c c c c c !{\color{LineGray}\vrule} c c c c c}
      \hline
      \rowcolor{HeaderGray}
       &
      \multicolumn{3}{c !{\color{LineGray}\vrule}}{Components} &
      \multicolumn{15}{c}{\textbf{CUHK-PEDES}} \\
      \cline{2-19}

      \rowcolor{HeaderGray}
       & & & &
      \multicolumn{5}{c !{\color{LineGray}\vrule}}{HR} &
      \multicolumn{5}{c !{\color{LineGray}\vrule}}{Ultra-LR} &
      \multicolumn{5}{c}{Mixed} \\
      \cline{5-19}

      \rowcolor{HeaderGray}
      \multirow{-3}{*}{No.} &
      \multirow{-2}{*}{RCR} & \multirow{-2}{*}{CR-RDA} & \multirow{-2}{*}{TGR} &
      R@1 & R@5 & R@10 & mAP & mINP &
      R@1 & R@5 & R@10 & mAP & mINP &
      R@1 & R@5 & R@10 & mAP & mINP \\
      \hline

      0 &  &  &  & 73.25 & 89.44 & 93.44 & 66.23 & 50.53 & 55.78 & 78.49 & 85.90 & 49.79 & 32.76 & 67.71 & 86.24 & 91.80 & 60.04 & 42.67 \\
      1 & \checkmark &  &  & 73.51 & 89.50 & 93.47 & 66.31 & 50.26 & 56.87 & 79.42 & 86.39 & 50.84 & 33.64 & 68.17 & 86.88 & 91.76 & 60.98 & 43.36 \\
      2 &  & \checkmark &  & 73.79 & 89.69 & 93.70 & 66.52 & \underline{50.73} & 57.46 & 79.19 & 86.19 & 52.51 & 33.63 & 68.38 & 86.62 & 91.42 & 61.80 & 43.96 \\
      3 &  &  & \checkmark & 73.31 & 89.33 & 93.43 & 66.23 & 50.59 & 57.88 & 80.18 & 86.69 & 51.27 & 34.39 & 69.07 & 86.95 & 91.67 & 61.34 & 43.94 \\
      4 & \checkmark & \checkmark &  & \underline{73.96} & \underline{89.72} & \underline{93.72} & \underline{66.67} & 50.55 & 59.60 & 80.95 & 86.90 & 52.72 & 34.49 & \underline{69.45} & 87.02 & 91.82 & 62.42 & 45.19 \\
      5 & \checkmark &  & \checkmark & 73.61 & 89.56 & 93.47 & 66.33 & 50.50 & \underline{60.89} & \underline{82.77} & \underline{87.89} & 53.81 & 35.40 & 69.43 & 87.31 & \underline{92.06} & 63.40 & 44.97 \\
      6 &  & \checkmark & \checkmark & 73.84 & 89.67 & 93.58 & 66.55 & 50.57 & 59.58 & 82.48 & 87.37 & \underline{53.82} & \underline{35.93} & 68.74 & \underline{87.91} & 91.75 & \underline{63.64} & \underline{45.38} \\
      \hline
      \rowcolor{OursBlue}
      7 & \checkmark & \checkmark & \checkmark & \textbf{74.10} & \textbf{89.84} & \textbf{93.78} & \textbf{66.68} & \textbf{50.63} & \textbf{61.81} & \textbf{83.59} & \textbf{88.08} & \textbf{55.84} & \textbf{36.32} & \textbf{70.36} & \textbf{88.09} & \textbf{91.90} & \textbf{64.05} & \textbf{46.40} \\
      \hline
    \end{tabular}
  }
  \captionsetup{justification=centering, singlelinecheck=false}
  \caption{\textbf{Component ablation.}
  Ablation study of different components on CUHK-PEDES with different resolutions.}
  \label{tab:ablation}
  \vspace{-2mm}
\end{table*}

\begin{figure}[t]
    \centering
    \includegraphics[width=0.45\textwidth]{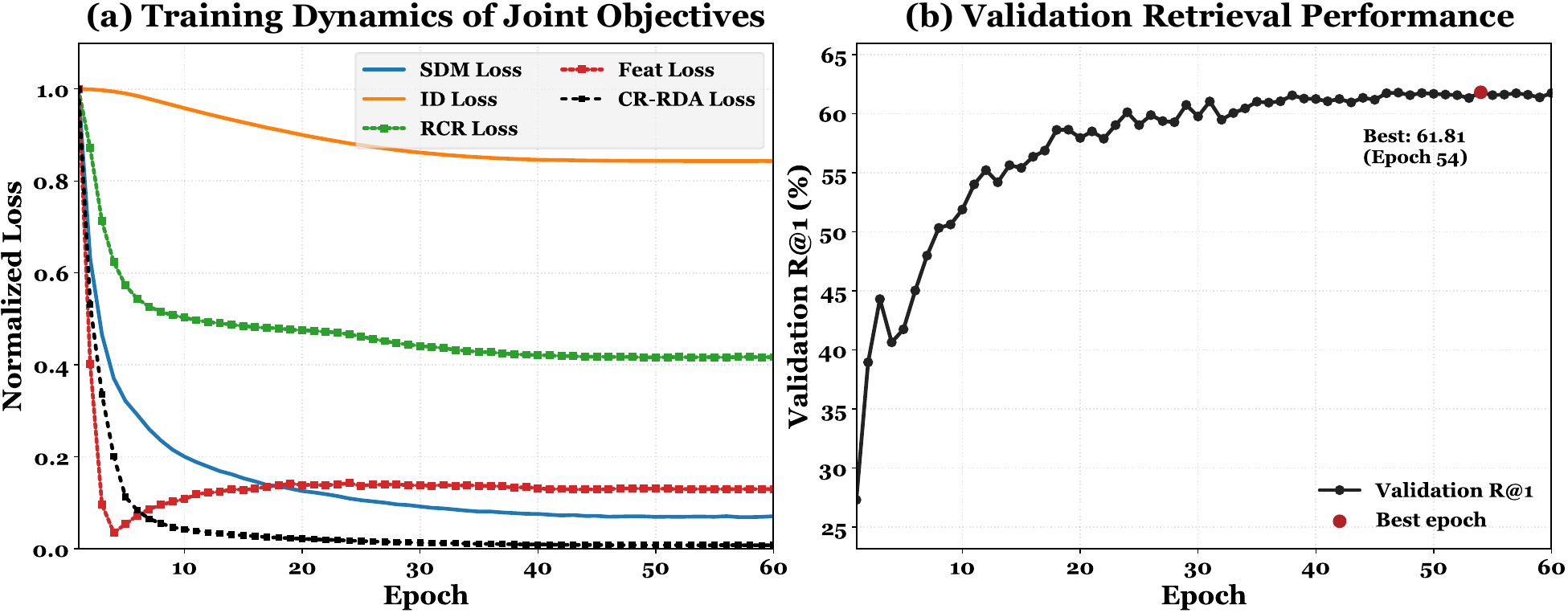}
    \caption{\textbf{Training dynamics under cross-resolution training.}
    (a) optimization curves of the total objective and its major components.
    (b) validation retrieval performance under Ultra-LR during training.
    The curves show stable joint optimization without obvious objective conflict.}
    \label{fig:train_dyn}
    \vspace{-4mm}
\end{figure}

\textbf{Optimization and Hyper-Parameters. }
We train the model for 60 epochs using Adam with batch size 64.
The base learning rate is $1\times10^{-5}$ for the pretrained CLIP encoders, while randomly initialized modules use a $5\times$ larger learning rate ($5\times10^{-5}$).
We adopt a cosine learning-rate schedule with 5 warm-up epochs and set the temperature to $\tau=0.02$.
Mini-batches are constructed with random sampling.
The Adam coefficients are set to $(0.9, 0.999)$, and the weight decay is $4\times10^{-5}$.

\textbf{Inference. }
During inference, the RCR, TGR, and CR-RDA modules are disabled, and the retrieval process is based solely on the similarity between visual and textual embeddings. This ensures that inference does not incur additional computational cost.


\textbf{Complexity Analysis. }
As shown in Table~\ref{tab:complexity}, CRST introduces a parameter increase over IRRA (\textit{194.5M vs. 196.7M}) while maintaining identical inference efficiency, indicating that our semantic transfer is lightweight.
Since RCR, TGR, and CR-RDA are training-time constraints, inference relies solely on the CLIP-style dual encoder, ensuring robustness without computational cost.

\subsection{Comparison with State-of-the-art Methods}
\textbf{Cross-Resolution Robustness. }
Table~\ref{tab:lr_all} evaluates Ultra-LR and Mixed settings, where resolution variance dominates surveillance deployments.
CRST improves robustness across three benchmarks under both severe degradation and heterogeneous galleries.
On CUHK-PEDES, CRST reaches 61.81/55.84 (R@1/mAP) in Ultra-LR and 70.36/64.05 in Mixed, outperforming {DM-Adapter\dag} (56.91/51.27 and 68.38/61.12) by 4.90/4.57 and 1.98/2.93.
On ICFG-PEDES, it achieves 46.17/25.94 in Ultra-LR and 58.11/34.45 in Mixed, improving over {DM-Adapter\dag} (41.87/21.52 and 56.64/31.51) by 4.30/4.42 and 1.47/2.94.
On RSTPReid, CRST attains 49.55/37.80 in Ultra-LR and 56.20/44.69 in Mixed, yielding +4.70/+3.54 and +1.13/+2.59 over {DM-Adapter\dag} (44.85/34.26 and 55.07/42.10).
Beyond the bicubic protocol, CRST remains robust under a coupled degradation setting with compression, blur, low-light, and noise: on CUHK-PEDES it achieves 57.02/51.27 in Ultra-LR and 67.08/61.11 in Mixed, outperforming the Baseline by 6.66/7.65 and 4.87/6.02 (Table~3 in Appendix), showing that the gain is not limited to clean synthetic downsampling.
Improvements correspond to Figure~\ref{fig:problem_show}: Ultra-LR triggers ERC, where unreliable evidence harms grounding, while Mixed exposes RDD, i.e., ranking drift under heterogeneous galleries.
Overall, CRST remains stronger than CLIP-based baselines retrained under the paired dual setting~\cite{bib52,bib53}, while its robustness beyond the protocol is further supported in Appendix~B--D.

\begin{figure}[t]
    \centering
    \includegraphics[width=0.45\textwidth]{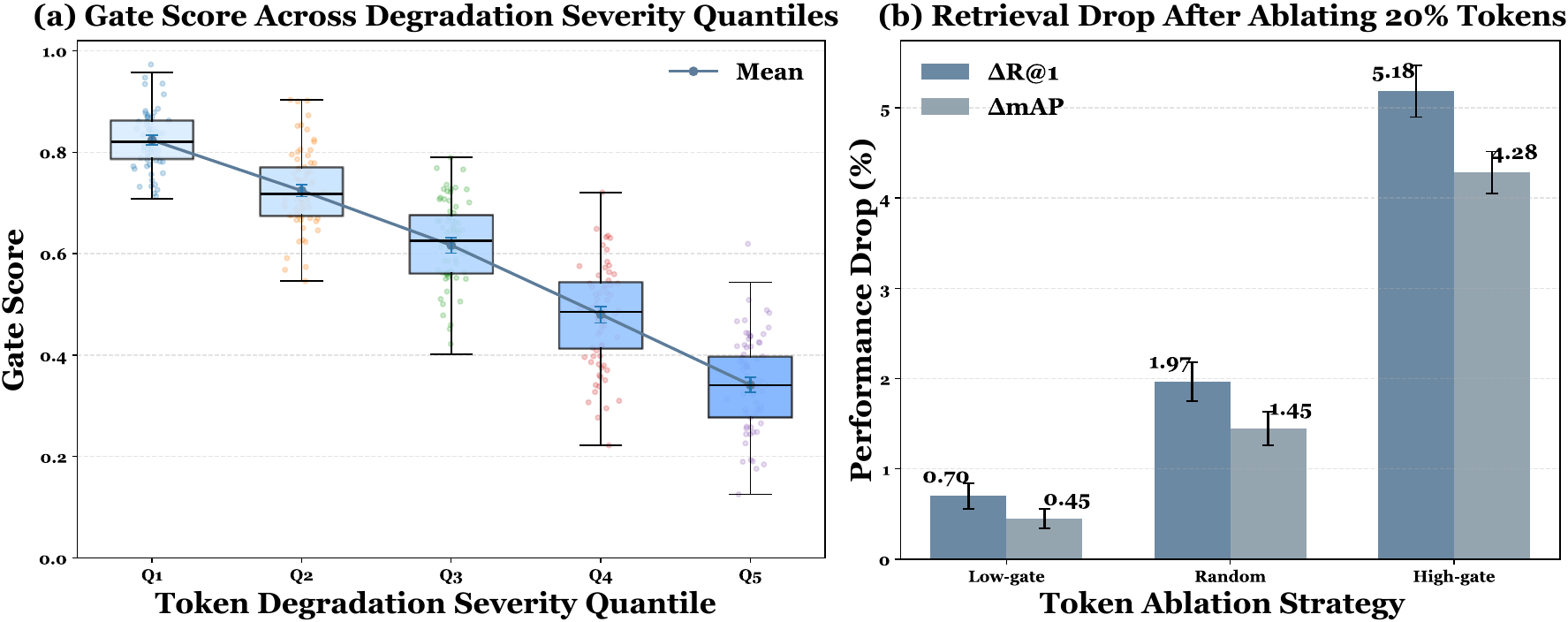}
    \caption{\textbf{Behavior of RCR gates under token degradation. }
    (a) Gate scores decrease consistently from Q1 to Q5, where tokens are grouped by increasing degradation severity, indicating lower reliability for more degraded tokens. 
    (b) Ablating high-gate tokens causes much larger drops in R@1 and mAP than ablating low-gate tokens, suggesting that the learned gates are functionally meaningful for retrieval.
    }
    \label{fig:rcr_vis}
    \vspace{-2mm}
\end{figure}

\textbf{Standard HR Results. }
Table~\ref{tab:hr_all} summarizes the comparison under the conventional HR protocol on CUHK-PEDES, ICFG-PEDES, and RSTPReid~\cite{bib50}.
CRST achieves the best results on CUHK-PEDES (74.10/66.68) and RSTPReid (61.05/47.94), while remaining competitive on ICFG-PEDES, where TBPS-CLIP reports 65.05/39.83 compared with our 64.71/38.75.
This verifies that introducing cross-resolution semantic transfer does not sacrifice the standard HR retrieval capability, while keeping the CLIP-style inference pipeline.

\begin{figure}[t]
    \centering
    \includegraphics[width=0.45\textwidth]{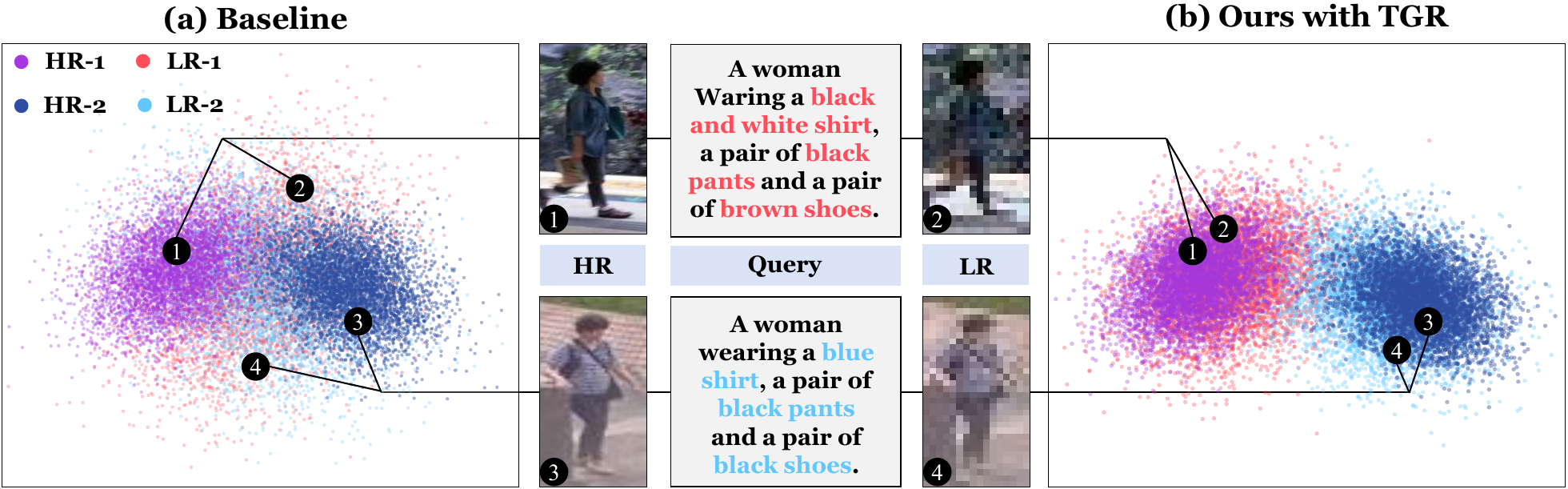}
    \caption{\textbf{Text-guided semantic recovery under UltraLR. }
    Left and right panels show baseline and TGR-refined t-SNE embeddings. Middle cards show two HR/LR cases. TGR pulls UltraLR samples back to the correct HR-consistent clusters.}
    \label{fig:tgr_vis}
    \vspace{-2mm}
\end{figure}

\textbf{Progressive Degradation Behavior. }
Figure~\ref{fig:rank_map_cmp} summarizes performance trends under progressive resolution degradation.
Across all benchmarks, CRST consistently outperforms the baseline without explicit cross-resolution modeling at every resolution level, and the margin becomes larger as the resolution decreases.
More importantly, CRST degrades more gracefully from HR to Ultra-LR, showing a noticeably smaller performance drop under severe degradation.
This trend confirms that CRST is substantially less sensitive to resolution disturbance: as visual details gradually vanish, our semantic transfer constraints help preserve reliable grounding and maintain a more stable similarity landscape, leading to more robust retrieval under LR and mixed-resolution conditions.

\begin{figure}[t]
\centering
\includegraphics[width=0.45\textwidth]{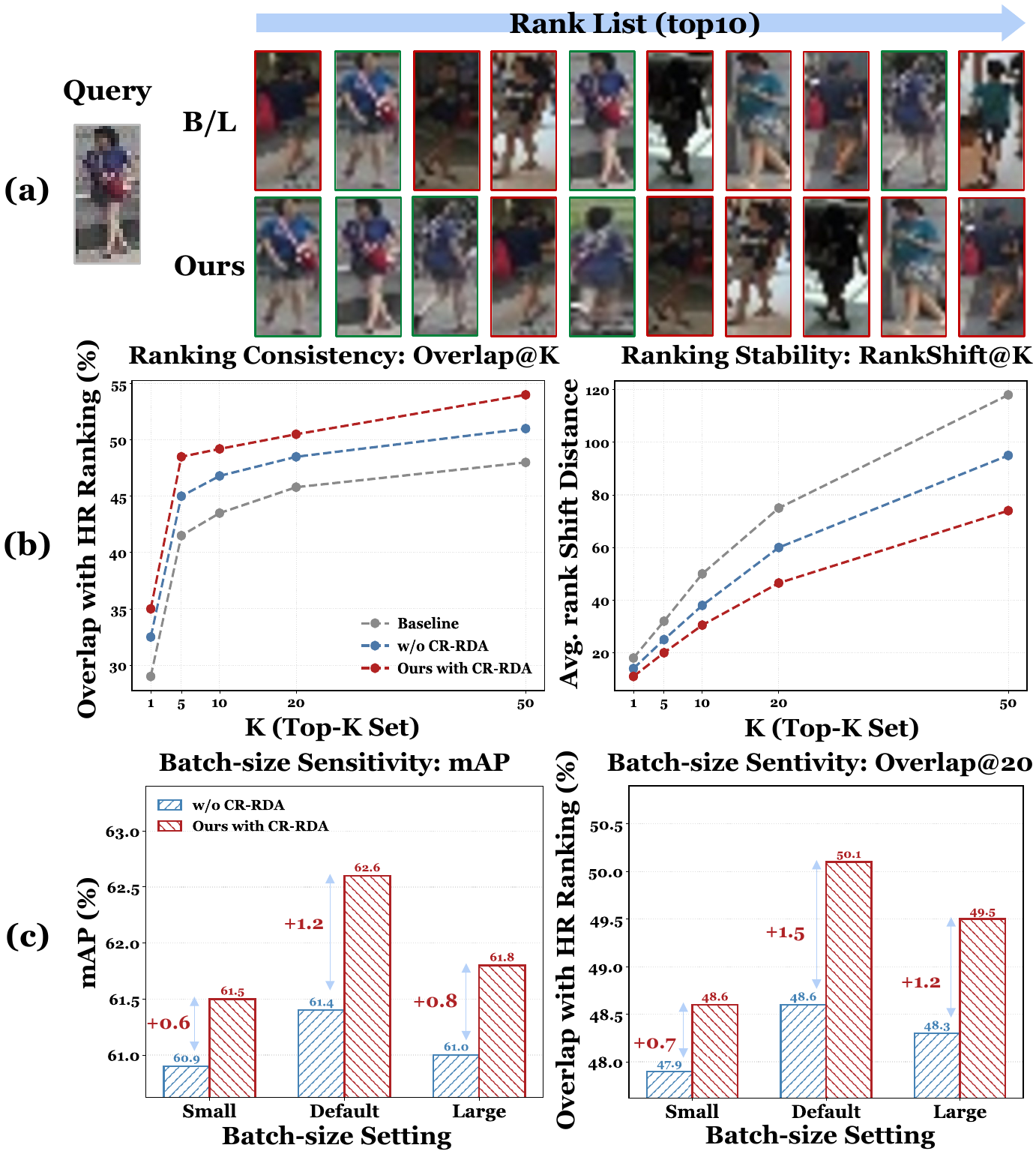}
\caption{\textbf{Effect of CR-RDA on retrieval ranking under Ultra-LR.}
(a) A representative Top-10 retrieval case (B/L: baseline; green: correct; red: incorrect match).
(b) HR-referenced ranking consistency and stability analysis.
(c) Batch-size sensitivity of CR-RDA.
}
\label{fig:crrda_evidence}
\vspace{-3mm}
\end{figure}

\subsection{Ablation Studies}\frenchspacing
We analyze the contribution of each component on CUHK-PEDES under three evaluation settings.
Table~\ref{tab:ablation} reports Rank-1/Rank-5/Rank-10, mAP and mINP~\cite{bib46}.

\textbf{Overall effectiveness of CRST. }
Starting from a baseline without explicit cross-resolution modeling, CRST improves performance, with the largest gains under severe resolution degradation.
On Ultra-LR, Rank-1 increases from 55.78 to 61.81 and mAP from 49.79 to 55.84, while on Mixed, Rank-1 rises from 67.71 to 70.36 and mAP from 60.04 to 64.05.
HR performance slightly improves from 73.25/66.23 to 74.10/66.68, suggesting modeling resolution disturbance does not sacrifice HR accuracy.

\textbf{Training dynamics and joint optimization. }
To verify that the objectives in CRST are jointly optimizable, Figure~\ref{fig:train_dyn} shows the training dynamics under cross-resolution training. 
The total loss and major auxiliary terms remain stable during optimization, while the validation performance improves progressively under Ultra-LR. 
This indicates that RCR, TGR, and CR-RDA act complementarily rather than causing persistent objective conflict. 
Combined with Table~\ref{tab:ablation}, these curves support that CRST benefits from stable joint optimization instead of heuristic loss stacking.

\textbf{RCR: Reliability-aware evidence gating. }
Under Ultra-LR, RCR improves R@1/mAP from 55.78/49.79 to 56.87/50.84. 
As shown in Figure~\ref{fig:rcr_vis}, gate scores decrease from Q1 to Q5 with increasing token degradation severity, and ablating high-gate tokens causes larger drops in R@1 and mAP than ablating low-gate tokens. 
These results suggest that RCR assigns lower reliability to more degraded tokens and that the gates are meaningful for retrieval.

\textbf{TGR: Text-guided semantic recovery. }
TGR improves Ultra-LR performance to 57.88/51.27 in R@1/mAP. Figure~\ref{fig:tgr_vis} shows that degraded samples are pulled back toward HR-consistent semantic clusters, while the representative cases indicate that this correction is guided by discriminative textual cues rather than merely producing tighter feature grouping.

\textbf{CR-RDA: HR-referenced ranking-geometry stabilization. }
CR-RDA improves Ultra-LR performance to 57.46/52.51 in R@1/mAP. As shown in Figure~\ref{fig:crrda_evidence}, it moves correct matches to earlier ranks, achieves higher Overlap@K and lower RankShift@K against the HR reference, and improves mAP and Overlap@20 across batch-size settings. These results indicate that CR-RDA preserves HR-consistent neighborhood structure, reduces near-neighbor rank swaps, and remains effective beyond a batch configuration.
Under Mixed resolution, CR-RDA regularizes hard-negative ordering and neighborhood consistency. Therefore, its benefit is often more visible in mAP than in R@1, since rank quality can improve even when the top-1 position changes marginally.

Combining components yields larger gains than any single one, and CRST performs best across all settings, verifying that token-level, feature-level, and distribution-level transfers are necessary to address ERC and RDD.

\begin{figure}
  \centering
  \includegraphics[width=0.47\textwidth]{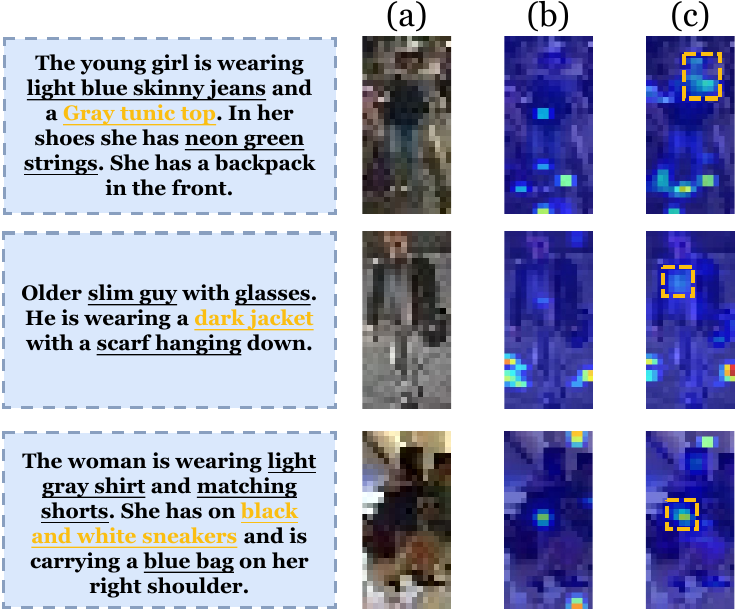}
  \caption{\textbf{Visualization of attention maps. }
  (a) Input images, (b) {Baseline}, (c) CRST.}
  \label{fig:heatmap}
\end{figure}

\subsection{Visualization}
Figure~\ref{fig:heatmap} visualizes word-to-region response maps under LR inputs.
Compared with baseline, CRST yields sharper and more localized activations on discriminative regions, with higher contrast between relevant evidence and background instead of diffuse responses.
This suggests that under severe resolution loss, CRST suppresses unreliable visual evidence and reduces spurious correlations, leading to more faithful word-region grounding.

\subsection{Limitations}
CRST targets resolution variance in surveillance TIPR and is evaluated under a controlled discrete-resolution protocol for cross-resolution study. Beyond this setting, Appendix~B--D verify robustness under realistic coupled degradations, noisy resolution labels, and an SR-aware alternative pipeline. Still, continuous or uncertainty-aware quality conditioning may further improve generalization~\cite{bib55}, and device-specific artifacts (\textit{e.g., motion blur, severe sensor noise}) as well as ambiguous queries remain challenging.

\section{Conclusion}
In this paper, we study cross-resolution text-to-image person retrieval and characterize its failures through two coupled issues: Evidence Reliability Collapse and Ranking Distribution Drift. These challenges arise because low-resolution degradation not only corrupts visual evidence for cross-modal grounding, but also reshapes similarity neighborhoods when images of different resolutions coexist in the gallery. To address them, we propose CRST, an HR-referenced semantic transfer framework for LR retrieval, which integrates reliability-aware evidence reasoning, prior-guided semantic correction, and hard-negative ordering consistency within a unified CLIP-style architecture. Experiments on CUHK-PEDES, ICFG-PEDES, and RSTPReid show that CRST improves robustness in Ultra-LR and Mixed settings without sacrificing HR accuracy.

\printbibliography

\end{document}